\documentclass[letterpaper, journal, twoside]{IEEEtran}

\usepackage{amsmath,amsfonts}
\usepackage{algorithmic}
\usepackage{array}
\usepackage{subfig}
\usepackage{textcomp}
\usepackage{stfloats}
\usepackage{url}
\usepackage{verbatim}
\usepackage{graphicx}
\usepackage{balance}

\usepackage{multirow}
\usepackage{booktabs}
\usepackage{xcolor}
\usepackage{colortbl}
\usepackage{hyperref}

\newcommand{\ours}[1]{TLC-Calib}

\newcommand{\best}[1]{\cellcolor{colorf}{#1}}
\newcommand{\second}[1]{\cellcolor{colors}{#1}}
\newcommand{\third}[1]{\cellcolor{colort}{#1}}
\newcommand{\bxref}[1]{\cellcolor{colorref}{#1}}

\definecolor{colorf}{HTML}{FFB2B2}
\definecolor{colors}{HTML}{FFD9B2}
\definecolor{colort}{HTML}{FFFFB2}
\definecolor{colorref}{HTML}{C8C8C8}

\newcommand{\boxcolorref}[1]{%
  \begingroup\setlength{\fboxsep}{1pt}%
  \colorbox{colorref}{\hspace*{2pt}\vphantom{Ay}#1\hspace*{2pt}}%
  \endgroup
}

\DeclareMathOperator*{\argmin}{arg\,min}

\begin{document}

\title{Targetless LiDAR-Camera Calibration with \\ Neural Gaussian Splatting}

\author{Haebeom Jung$^{1}$,
Namtae Kim$^{1}$,
Jungwoo Kim$^{2}$,
and Jaesik Park$^{1*}$,~\IEEEmembership{Member,~IEEE}
\thanks{Manuscript received: October 8, 2025; Revised December 18, 2025; Accepted January 13, 2026.
%
This paper was recommended for publication by Editor Pascal Vasseur upon evaluation of the Associate Editor and Reviewers comments.
This work was supported by IITP Grant (RS-2021-II211343: AI Graduate School Program at Seoul National University (5\%) and RS-2023-00227993: Detailed 3D reconstruction for
urban areas from unstructured images (60\%)) and NRF Grant (No.2023R1A1C200781211 (35\%)) funded by the Korea Government (MSIT).}%
\thanks{$^{1}$H.~Jung, N.~Kim, and J.~Park are with the Department of Interdisciplinary Program in Artificial Intelligence, Seoul National University, Seoul 08826, South Korea (e-mail: \texttt{\{haebeom.jung, knt0613, jaesik.park\}@snu.ac.kr})}%
\thanks{$^{2}$J.~Kim is with the Department of Artificial Intelligence, Yonsei University, Seoul 03722, South Korea (e-mail: \texttt{jungwkim@yonsei.ac.kr})}%
\thanks{$^*$Jaesik Park is the corresponding author of this work.}
\thanks{Digital Object Identifier (DOI): see top of this page.}
}

\markboth{IEEE Robotics and Automation Letters. Preprint Version. Accepted January, 2026}
{Jung \MakeLowercase{et al.}: Targetless LiDAR-Camera Calibration with Neural Gaussian Splatting} 

\maketitle

\begin{abstract}
Accurate LiDAR-camera calibration is crucial for multi-sensor systems. However, traditional methods often rely on physical targets, which are impractical for real-world deployment. 
Moreover, even carefully calibrated extrinsics can degrade over time due to sensor drift or external disturbances, necessitating periodic recalibration.
To address these challenges, we present a Targetless LiDAR–Camera Calibration (TLC-Calib) that jointly optimizes sensor poses with a neural Gaussian–based scene representation. 
Reliable LiDAR points are frozen as anchor Gaussians to preserve global structure, while auxiliary Gaussians prevent local overfitting under noisy initialization.
Our fully differentiable pipeline with photometric and geometric regularization achieves robust and generalizable calibration, consistently outperforming existing targetless methods on the \textsc{KITTI-360}, \textsc{Waymo}, and \textsc{Fast-LIVO2} datasets.
In addition, it yields more consistent Novel View Synthesis results, reflecting improved extrinsic alignment.
The project page is available at: \href{https://www.haebeom.com/tlc-calib-site/}{https://www.haebeom.com/tlc-calib-site/}.
\end{abstract}



\IEEEpeerreviewmaketitle


\section{Introduction}
\label{sec:intro}

\IEEEPARstart{R}{ecent} advances in Novel View Synthesis (NVS) have enabled increasingly sophisticated reconstruction of 3D scenes from 2D images~\cite{levoy1996light, gortler1996lumigraph,mildenhall2020nerf,kerbl20233dgs}.
Despite these innovations, achieving higher rendering quality and precise 3D geometry often requires accurate geometry using multi-sensor fusion, such as the integration of LiDAR and multiple cameras.
This complementary fusion provides richer and more accurate spatial information, and recent studies~\cite{zhao2024tclc, chen2025omnire} have reported substantial performance gains, especially in NVS tasks.

However, neural rendering techniques in multi-sensor setups rely heavily on accurate knowledge of each sensor’s mounting position and orientation, known as sensor extrinsics.
These parameters are not necessarily static.
Over time, even slight mechanical vibrations, thermal expansion, or physical impacts can induce subtle shifts in sensor positioning, resulting in misalignment and the need for periodic recalibration.

Target-based calibration methods~\cite{zhang2004extrinsic, geiger2012automatic, mirzaei20123d} are widely adopted as a standard solution.
For instance, placing checkerboard patterns or spherical reflectors within the shared field of view enables accurate pose estimation.
While effective, this approach can require costly infrastructure or large-scale target installations, especially in systems with multiple sensors or wide baselines. 
Moreover, even carefully calibrated target-based methods often struggle to align LiDAR and camera data at far distances, limiting their utility in real-world scenarios.

By contrast, targetless methods~\cite{levinson2013automatic, scaramuzza2007extrinsic, munoz2020targetless} calibrate sensors using only raw sensor data, leveraging environmental features such as planes or edges~\cite{munoz2020targetless}.
These methods eliminate the need for physical targets.
Nevertheless, they face significant challenges due to the intrinsic differences between LiDAR and camera modalities, particularly the sparsity of LiDAR point clouds.
Deep learning-based approaches~\cite{schneider2017regnet, lv2021lccnet} attempt to bridge this gap.
However, such approaches typically require large labeled datasets and often struggle to generalize to new sensor configurations or scenes.
%
%
Although NeRF-based methods~\cite{zhou2023inf, herau2023moisst, yang2024unical} can jointly optimize scene representations and sensor poses, their implicit volumetric nature results in high computational overhead, often scaling with the number of images.

In contrast, we employ a neural Gaussian representation to enable efficient and scalable optimization.
We propose \textit{\ours{}}, a targetless LiDAR-camera calibration framework built upon this representation.
By leveraging differentiable rendering, our method jointly optimizes sensor extrinsics and the scene representation without relying on explicit calibration targets or external supervision.
Joint optimization is essential, as pose estimation is tightly coupled with the underlying scene representation.
Recent studies~\cite{brachmann2024scene} show that even small pose inaccuracies can severely degrade NVS quality, highlighting the importance of precise camera calibration.
To support scalable optimization across diverse scenes, we introduce adaptive voxel control, which automatically adjusts anchor density based on scene scale and motion, eliminating the need for manual voxel resolution tuning.
Reliable LiDAR points are designated as anchor Gaussians to preserve global structure, while auxiliary Gaussians provide local flexibility and mitigate overfitting under inaccurate initial poses.
This design enables robust optimization even with noisy initialization and improves alignment quality across diverse environments.

In summary, the primary contributions of this paper are as follows:
(i) We ensure metric scene scale by designating reliable LiDAR points as anchor Gaussians to preserve overall scene structure, while auxiliary Gaussians regularize local geometry under challenging initialization.
(ii) We integrate adaptive voxel control and Gaussian scale regularization to reduce redundant anchor Gaussians and suppress over-dominant anisotropic Gaussians that hinder optimization stability.
(iii) We validate our approach on three real-world setups, including two autonomous driving datasets and a handheld solid-state LiDAR setup, demonstrating strong generalization, high calibration accuracy, and rendering quality.

    
    

\section{Related Work}
\label{sec:related_work}


\subsection{Targetless Sensor Calibration}
Targetless calibration methods align sensors using environmental cues instead of physical markers.
Edge-based approaches~\cite{levinson2013automatic, zhang2021line} extract geometric edges from point clouds and images to estimate sensor extrinsics.
In parallel, learning-based approaches have also been actively studied.
RegNet~\cite{schneider2017regnet} employs convolutional neural networks to predict extrinsic parameters between LiDAR scans and images, while LCCNet~\cite{lv2021lccnet} improves upon this by introducing a cost volume for more robust estimation.
Additionally, segmentation-based methods maximize overlap regions~\cite{zhu2020online}, align object edges~\cite{rotter2022automatic}, or leverage SAM-based masks for calibration~\cite{luo2023calibanything}.
However, the accuracy of such methods is often limited by the quality of segmentation.

\subsection{Neural Rendering for Sensor Calibration}

Neural rendering methods such as NeRF~\cite{mildenhall2020nerf} and 3DGS~\cite{kerbl20233dgs} have been extended to jointly refine camera poses and scene geometry via photometric loss~\cite{herau20243dgscalib, zhou2023inf, herau2023moisst, yang2024unical}.
However, these approaches are primarily designed for camera-only systems and are difficult to extend to multi-sensor settings due to scale ambiguity and modality gaps between LiDAR and camera data.
Recently, neural rendering has also been explored for LiDAR-camera extrinsic calibration.
Early NeRF-based methods~\cite{zhou2023inf,herau2023moisst,yang2024unical} formulate calibration as a radiance field optimization problem, but their high computational cost limits practical applicability.
To alleviate this limitation, recent works~\cite{herau20243dgscalib, zhou2025robust} adopt 3DGS to accelerate optimization.
3DGS-Calib~\cite{herau20243dgscalib} fixes Gaussians on LiDAR points to guide calibration, but its reliance on hash-grid encodings makes the optimization sensitive to scene complexity and hyperparameter choices.
RobustCalib~\cite{zhou2025robust} introduces a two-stage strategy that learns geometric constraints from LiDAR point clouds using 2DGS, followed by extrinsic calibration with reprojection and triangulation losses.
However, its performance depends on the quality of the estimated surface normals, which may degrade under sparse LiDAR observations.
In contrast, \ours{} introduces anchor and auxiliary Gaussians to construct a fully differentiable scene representation that extends beyond LiDAR-overlapped regions, enabling robust and generalizable calibration across diverse environments.

\section{Method}
\label{sec:method}

\subsection{Preliminary: 3DGS with Differentiable Pose Rasterization}

We employ 3D Gaussian Splatting (3DGS) as a differentiable scene representation for jointly optimizing the scene and camera poses.
The scene is represented as a set of anisotropic 3D Gaussians, each parameterized by a center $\boldsymbol{\mu}_i \in \mathbb{R}^3$, covariance $\mathbf{\Sigma}_i$, opacity $\alpha_i$, and view-dependent color coefficients $\mathbf{c}_i$.
We consider a multi-camera setup with $C$ cameras, indexed by $c \in \{1, \dots, C\}$.
Given the camera intrinsics $\mathbf{K}$ and camera pose $\mathbf{T}^c \in \mathrm{SE}(3)$, each Gaussian is projected onto the image plane as a 2D Gaussian $\mathbf{G}_i^{2D}$~\cite{zwicker2002ewa}.

After sorting Gaussians in a front-to-back order along the viewing direction, the rendered color at pixel $\mathbf{u}$ is obtained via alpha compositing:
\begin{equation}
\mathbf{C}(\mathbf{u}) = \sum_{i=1}^{N} 
\mathbf{c}_i \alpha_i \mathbf{G}^{\text{2D}}_i(\mathbf{u})
\prod_{j=1}^{i-1} \left(1 - \alpha_j \mathbf{G}^{\text{2D}}_j(\mathbf{u})\right).
\label{eq:color_term}
\end{equation}

The rendered color in Eq.~\eqref{eq:color_term} is an explicit function of the camera pose through the projection of Gaussian means and view-dependent colors.
This allows gradients of a photometric loss $\mathcal{L}$ to be analytically propagated to the camera pose, following the pose-differentiable rasterization framework of Gaussian Splatting SLAM~\cite{matsuki2024monogs}.
Using the chain rule, the gradient of the loss with respect to the camera pose is expressed as
\begin{equation}
\begin{aligned}
\frac{\partial \mathcal{L}}{\partial \mathbf{T}^c} =
\sum_i \Big(
    \frac{\partial \mathcal{L}}{\partial \boldsymbol{\mu}_i^{\text{2D}}}
    \frac{\partial \boldsymbol{\mu}_i^{\text{2D}}}{\partial \boldsymbol{\mu}_i^c}
    \frac{\partial \boldsymbol{\mu}_i^c}{\partial \mathbf{T}^c}
    + \frac{\partial \mathcal{L}}{\partial \mathbf{c}_i}
    \frac{\partial \mathbf{c}_i}{\partial \mathbf{T}^c}
\Big),
\end{aligned}
\label{eq:photo_loss}
\end{equation}
where $\boldsymbol{\mu}_i^{\text{2D}}$ denotes the projected mean.
All required Jacobians follow standard rigid-body transformation rules on $\mathrm{SE}(3)$.

Following~\cite{matsuki2024monogs}, the camera pose is updated directly on the Lie group as
\begin{equation}
\mathbf{T}^c \leftarrow \exp\!\left(-\lambda \frac{\partial \mathcal{L}}{\partial \mathbf{T}^c}\right)\mathbf{T}^c,
\end{equation}
ensuring geometrically consistent and fully differentiable pose optimization.

\begin{figure}[t]
    \centering
    \captionsetup{font=footnotesize}
    \includegraphics[width=1.0\linewidth]{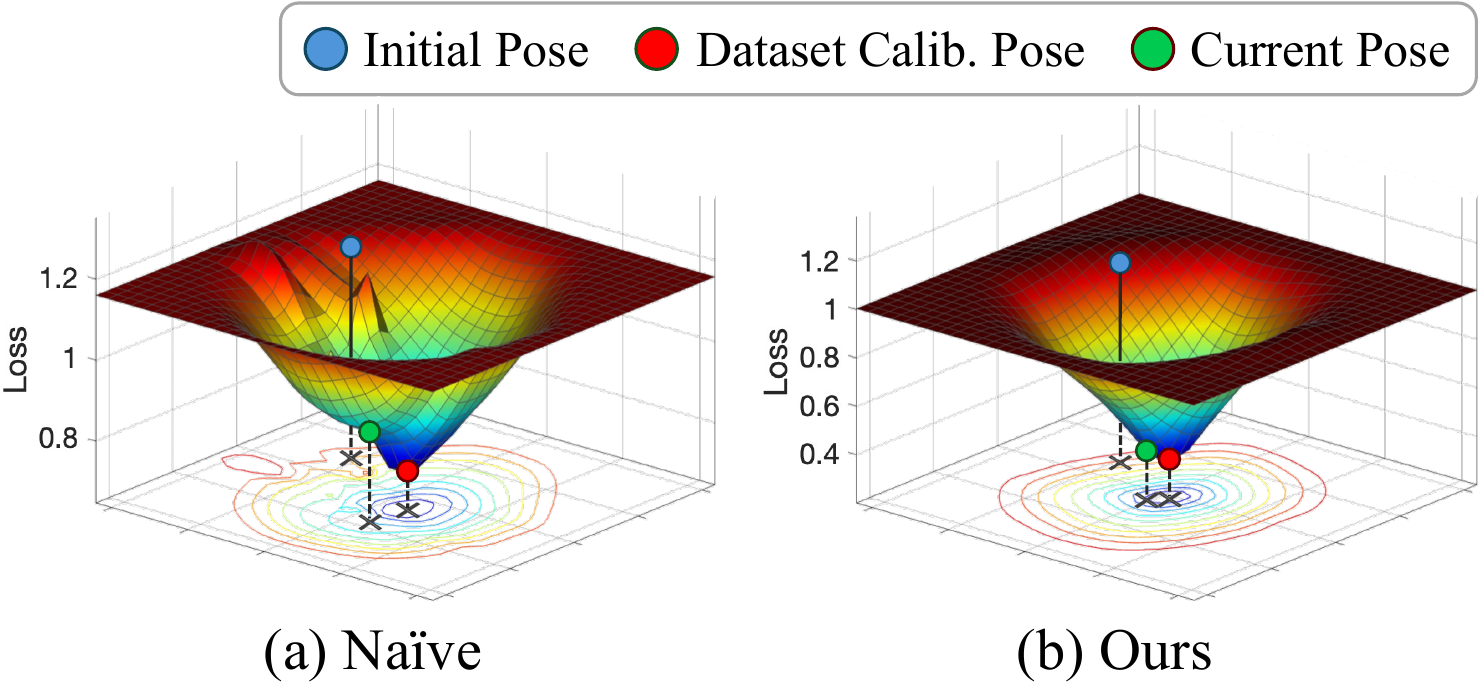}
    \caption{
        An empirical example of optimization landscapes.
        We construct the loss surface by sampling pose perturbations around the dataset calibration and evaluating the photometric loss of rendered views.
        (a) The na\"{\i}ve baseline (3DGS~\cite{kerbl20233dgs} + rig optimization) overfits individual views, yielding an irregular landscape where the pose becomes trapped in local minima and fails to reach the dataset calibration.
        (b) Our method mitigates view-dependent overfitting using neural Gaussians, producing a smoother loss surface and more stable convergence toward the dataset calibration.
    }
    \label{fig:loss_landscape}    
    \vspace{-3mm}
\end{figure}
\begin{figure*}[t]
    \centering
    \captionsetup{font=footnotesize}
    \includegraphics[width=0.99\linewidth]{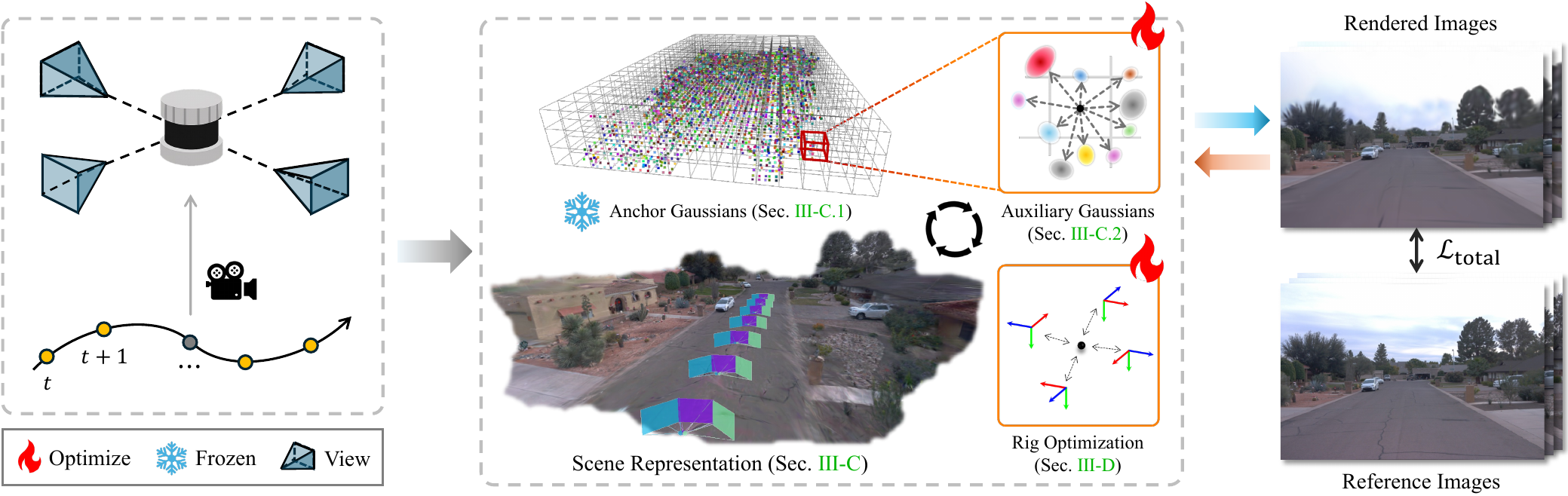}
    \caption{
        Overview of the \ours{} pipeline.
        After aggregating LiDAR scans into a globally aligned point cloud, anchor Gaussians serve as fixed geometric references (their positions are not optimized), while auxiliary Gaussians adapt to local geometry and guide extrinsic optimization through photometric loss.
        Unlike anchor Gaussians, auxiliary Gaussians serve as learnable buffers around anchors, helping the optimization avoid local minima.
        Additionally, the camera rig optimization strategy jointly refines all cameras with respect to the scene, ensuring consistent and stable calibration across views.
    }
    \label{fig:overview}
    \vspace{-2mm}
\end{figure*}

\subsection{Overview}

We propose a neural Gaussian-based approach for targetless calibration in a LiDAR and multi-camera setup.
We adopt the LiDAR as the reference sensor, treating its coordinate frame as the global reference and calibrating all cameras relative to it.
%
This design choice is driven by the wide horizontal field of view and precise 3D geometry of LiDAR, which enables reliable odometry estimation through SLAM~\cite{chen2024ig}, ICP-based registration~\cite{vizzo2023kiss}, or fusion with GPS and IMU for drift-free trajectories.
Such multi-sensor fusion pipelines are widely adopted in autonomous driving datasets, including \textsc{KITTI-360}~\cite{liao2022kitti360} and \textsc{Waymo}~\cite{sun2020waymo}.
Building on this, we aggregate LiDAR point clouds over time using odometry, ensuring geometric consistency before calibration.
In this paper, we assume that LiDAR poses are given and that sensor timestamps are well synchronized.
An overview of the proposed pipeline is shown in Fig.~\ref{fig:overview}.

\subsection{Neural Scene Representation}
\label{sec:scene_representation}

\subsubsection{Anchor Gaussians}
\label{sec:anchor_gaussians}
Since LiDAR serves as the reference sensor, we aggregate point clouds across timestamps
$t \in \{1, \dots, T\}$ to form $\mathcal{P} = \bigcup_{t=1}^T \mathcal{P}_t$, where $\mathcal{P}_t$ denotes each LiDAR scan.
To control point density, we voxelize $\mathcal{P}$ with an adaptively determined voxel size $\varepsilon^*$ (Sec.~\ref{sec:adaptive_voxel_control}) and select a subset of representative points.
Voxelization is used solely for downsampling and spatial indexing.
Anchor Gaussians are instantiated at the original coordinates of the selected LiDAR points, without shifting them to voxel centers or grid-aligned locations.
Each selected LiDAR point $\mathbf{p}_j \in \mathcal{P}$ directly defines an anchor Gaussian with center $\mathbf{v}_i = \mathbf{p}_j$, providing a stable reference in real-world coordinates.
Anchor positions remain fixed throughout training to preserve global scale and mitigate drift in LiDAR-camera calibration.
In addition, anchors with persistently low opacity are treated as floaters and removed during training.

\subsubsection{Auxiliary Gaussians}
\label{sec:auxiliary_gaussians}
While anchor Gaussians remain static, we introduce auxiliary Gaussians to refine local geometry and improve pose convergence.
Following~\cite{lu2024scaffold}, each anchor Gaussian $\mathbf{v}_i$ is associated with a learned feature vector $\mathbf{f_i}$ that encodes local geometric context.
For each camera $c$, we use a view-dependent input $\mathbf{d}_{i,c}$, defined as the normalized viewing direction from the anchor to the camera center.
The scalar $\ell_i$ denotes a learned scale parameter of the anchor Gaussian.
For each anchor $\mathbf{v}_i$, a lightweight MLP $\mathrm{F}_{\text{auxiliary}}$ predicts a fixed set of positional offsets $\boldsymbol{\delta}_i = \{\boldsymbol{\delta}_{i,1}, \dots, \boldsymbol{\delta}_{i,K}\}$, where $K$ is the number of auxiliary Gaussians per anchor.
The center of each auxiliary Gaussian, $\mathbf{m}_{i,k}$, is obtained by adding its offset to the corresponding anchor position:
\begin{equation}
    \boldsymbol{\delta}_i = \mathrm{F}_{\text{auxiliary}}(\mathbf{f_i},\, \mathbf{d}_{i,c},\, \ell_i),
    \quad
    \mathbf{m}_{i,k} = \mathbf{v}_i + \boldsymbol{\delta}_{i,k}.
\label{eq:anchor_position}
\end{equation}

Other Gaussian attributes such as covariance $\mathbf{\Sigma}_{i,k}$, color $\mathbf{c}_{i,k}$, and opacity $\boldsymbol{\alpha}_{i,k}$ are decoded via separate MLPs conditioned on $\{\mathbf{f_i}, \mathbf{d}_{i,c}, \ell_i\}$.



\subsubsection{Role of Auxiliary Gaussians}
\label{sec:role_of_aux_gaussians}
\begin{figure}[t]
    \centering
    \captionsetup{font=footnotesize}
    \includegraphics[width=1.0\linewidth]{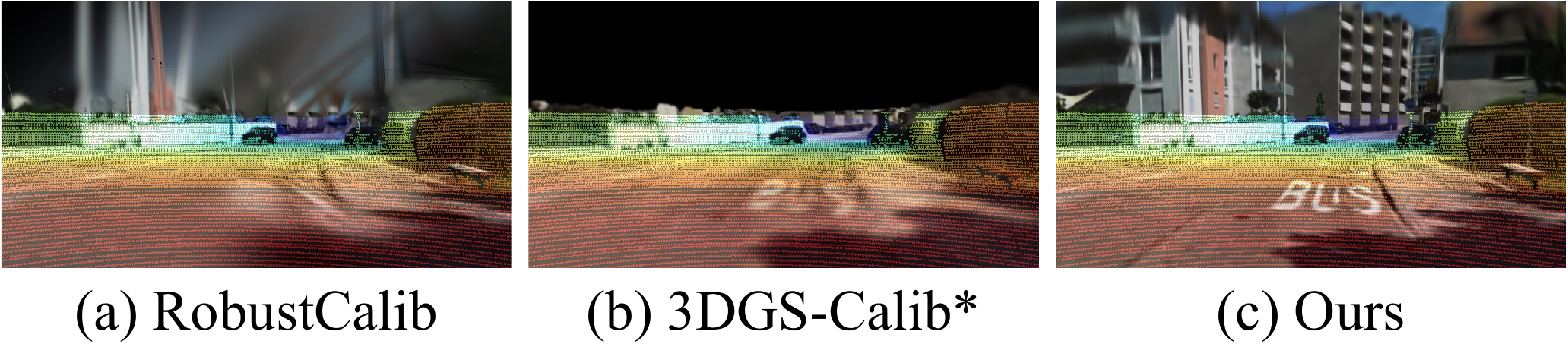}  
    \caption{
        Rendering results of (a)-(c), with the initial LiDAR points overlaid.
        Methods in (a) and (b) focus on LiDAR-observed regions with fixed geometry and consequently omit photometrically informative areas.
        In contrast, our method explicitly represents camera-observed regions beyond LiDAR coverage.
    }
    \label{fig:scene_rep_comp}    
    \vspace{-3mm}
\end{figure}

Auxiliary Gaussians provide local, learnable support around anchors, allowing geometry and appearance to adjust during pose optimization.
They further enable gradient propagation in sparse or LiDAR-unobserved regions, such as the sky or upper building areas, by introducing trainable structures beyond LiDAR coverage.
This mechanism distinguishes our approach from prior methods~\cite{herau20243dgscalib, zhou2025robust}, which regresses Gaussian attributes directly from LiDAR points, thereby constraining the optimization to LiDAR-observed regions (see Fig.~\ref{fig:scene_rep_comp} for comparison).
Because the rendering loss cannot propagate to areas beyond LiDAR coverage, these methods discard photometrically informative regions.
In contrast, our auxiliary Gaussians expand spatial coverage while preserving global scale consistency, as they are derived from anchor features and absorb supervision from nearby pixels even in LiDAR-unobserved areas.
The effectiveness of this scheme is empirically validated in the ablation study in Sec.~\ref{sec:effect_mask}.


\begin{table*}[t!]
    \centering
    \captionsetup{font=footnotesize}
    \caption{
        Baseline comparison on \textsc{KITTI-360}~\cite{liao2022kitti360}.
        Calibration performance is measured by success rate (SR, \%), rotation error, and translation error across motion scenarios.
        SR denotes cameras within 1$^\circ$ and 20\,cm.
        Errors are averaged over 10 runs with color-coded rankings.
    }
    \resizebox{1.0\textwidth}{!}{
    \setlength{\tabcolsep}{1pt}
    \begin{tabular}{l|*3{c}|*3{c}|*3{c}|*3{c}|*3{c}}
    \toprule[1.0pt]
    \multirow{2}{*}{\small{Scenes}}
    & \multicolumn{3}{c|}{\small{CalibAnything~\cite{luo2023calibanything}}}
    & \multicolumn{3}{c|}{\small{INF~\cite{zhou2023inf}}}
    & \multicolumn{3}{c|}{\small{RobustCalib~\cite{zhou2025robust}}}
    & \multicolumn{3}{c|}{\small{3DGS-Calib$^*$~\cite{herau20243dgscalib}}}
    & \multicolumn{3}{c}{\small{Ours}} \\
    
    & {SR$\uparrow$} & $\mathbf{R}$($^\circ$)$\downarrow$ & $\mathbf{t}$(cm)$\downarrow$
    & {SR$\uparrow$} & $\mathbf{R}$($^\circ$)$\downarrow$ & $\mathbf{t}$(cm)$\downarrow$
    & {SR$\uparrow$} & $\mathbf{R}$($^\circ$)$\downarrow$ & $\mathbf{t}$(cm)$\downarrow$
    & {SR$\uparrow$} & $\mathbf{R}$($^\circ$)$\downarrow$ & $\mathbf{t}$(cm)$\downarrow$
    & {SR$\uparrow$} & $\mathbf{R}$($^\circ$)$\downarrow$ & $\mathbf{t}$(cm)$\downarrow$ \\
    
    \midrule
    \texttt{Straight} & { 50.0 } & { 2.01$\pm$1.54 } & { 39.9$\pm$34.8 } & { {100.} } & { \third{0.22$\pm$0.12} } & { \best{10.9$\pm$1.44} } & { {100.} } & { \second{0.21$\pm$0.13} } & { \second{12.3$\pm$3.16} } & { {75.0} } & { {0.84$\pm$0.32} } & { {15.7$\pm$5.65} } & { {100.} } & { \best{0.11$\pm$0.04} } & { \third{12.7$\pm$1.43} } \\
    
    \texttt{Small zigzag} & { 50.0 } & { 1.74$\pm$1.40 } & { \third{16.6$\pm$12.0} } & { {50.0} } & { \third{0.82$\pm$0.65} } & { {56.9$\pm$54.7} } & { {100.} } & { \second{0.44$\pm$0.12} } & { \best{9.53$\pm$3.35} } & { {50.0} } & { {0.84$\pm$0.80} } & { {18.1$\pm$8.86} } & { {100.} } & { \best{0.15$\pm$0.02} } & { \second{10.1$\pm$0.73} } \\
    
    \texttt{Large zigzag} & { 0.00 } & { 4.61$\pm$1.59 } & { 78.5$\pm$21.4 } & { {100.} } & { \second{0.20$\pm$0.10} } & { \second{10.4$\pm$1.83} } & { {92.5} } & { \third{0.70$\pm$2.39} } & { \third{11.6$\pm$13.1} } & { {0.00} } & { {2.36$\pm$0.50} } & { {47.1$\pm$19.0} } & { {100.} } & { \best{0.09$\pm$0.03} } & { \best{6.17$\pm$2.27} } \\
    
    \texttt{Small rotation} & { 0.00 } & { 9.29$\pm$2.62 } & { 92.3$\pm$15.4 } & { {75.0} } & { \second{0.23$\pm$0.11} } & { \second{17.4$\pm$3.54} } & { {7.50} } & { {3.45$\pm$6.15} } & { {106.$\pm$105.} } & { {75.0} } & { \third{1.84$\pm$2.27} } & { \third{18.7$\pm$10.9} } & { {100.} } & { \best{0.21$\pm$0.03} } & { \best{6.21$\pm$1.67} } \\
    
    \texttt{Large rotation} & { 37.5 } & { 1.88$\pm$1.35 } & { \third{55.6$\pm$49.3} } & { {47.5} } & { \second{0.57$\pm$0.47} } & { {67.4$\pm$51.0} } & { {35.0} } & { {11.5$\pm$20.8} } & { {58.0$\pm$44.2} } & { {50.0} } & { \third{0.94$\pm$0.32} } & { \second{18.4$\pm$1.99} } & { {100.} } & { \best{0.09$\pm$0.03} } & { \best{9.18$\pm$1.04} } \\
    
    \midrule
    Mean & { 27.5 } & { 3.91$\pm$3.39 } & { 56.6$\pm$40.3 } & { {74.5} } & { \second{0.41$\pm$0.45} } & { \third{32.6$\pm$41.5} } & { 67.0 } & { 3.26$\pm$10.6 } & { 39.4$\pm$63.7 } & { {50.0} } & { \third{1.36$\pm$1.28} } & { \second{23.6$\pm$16.0} } & { {100.} } & { \best{0.13$\pm$0.05} } & { \best{8.86$\pm$2.90} } \\
    
    \bottomrule[1.0pt]
    \end{tabular}}
    \label{tab:kitti360_acc}
    \vspace{-2mm}
\end{table*}
\begin{table}[t]
    \centering
    \captionsetup{font=footnotesize}
    \caption{
        Training time comparison with baseline methods.
    }
    \resizebox{0.92\columnwidth}{!}{
    \setlength{\tabcolsep}{2pt}
    \begin{tabular}{l|c|c|c|c|c}
    \toprule[1.0pt]
    & CalibAnything & INF & RobustCalib & 3DGS-Calib$^*$ & Ours \\
    
    \midrule
    {Time$\downarrow$} & {\textgreater~5h} & {\textgreater~4h} & {\textgreater~1h} & {\mbox{$\sim$~0.15h}} & {\mbox{$\sim$~0.18h}} \\
    
    \bottomrule[1.0pt]
    \end{tabular}}    
    \label{tab:time_comparison}
    \vspace{-3mm}
\end{table}


\subsection{Joint Optimization of Scene and Extrinsics}
\label{sec:joint_optimization}
We denote the LiDAR-to-camera extrinsic for camera $c$ as $\mathbf{T}_c^e$.
Given the scene representation described previously, we jointly optimize the 3D Gaussians $\mathbf{G}$ and the extrinsic parameters $\{\mathbf{T}_c^e\}_{c=1}^C$ corresponding to each of the $C$ cameras.

Formally, the optimization objective is:
\begin{equation}
    \min_{\mathbf{G}, \mathbf{T}_{c}^e}
    \sum_{c=1}^{C}\sum_{t=1}^{T}
    \mathcal{L}_{\text{total}}\Bigl(I'_{c,t}, I_{c,t};\,\mathbf{G},\mathbf{T}_{c}^{e}\Bigr),
\end{equation}
where $I'_{c,t}$ is the rendered image (as in Eq.~\ref{eq:color_term}), and each camera $c$ provides $T$ observed images.

In practice, we adopt a \textit{camera rig optimization} strategy based on per-image sequential updates.
At each iteration, a single training image $(c,t)$ is randomly sampled and rendered to compute its photometric loss.
The gradient computed from this image is then immediately applied to the shared extrinsic $\mathbf{T}_c^e$ of camera $c$:
\begin{equation}
    \mathbf{T}_c^e \leftarrow 
    \mathbf{T}_c^e 
    - \alpha \nabla_{\mathbf{T}_c^e}
    \mathcal{L}_{\text{photo}}\bigl(I'_{c,t}, I_{c,t}\bigr),
\end{equation}
where $\alpha$ is the step size and $\mathcal{L}_{\text{photo}}$ denotes the pose-differentiable photometric loss (Eq.~\ref{eq:photo_loss}).
Because all frames captured by camera $c$ share a common extrinsic, the updated $\mathbf{T}_c^e$ is immediately applied to all images from that camera.
This per-view update allows each observation to directly correct pose misalignments, without requiring gradient accumulation across multiple views.

To analyze the behavior of the joint optimization, we examine how different scene representations shape the underlying energy landscape.
We measure the photometric loss surface by sampling pose perturbations around the dataset calibration and evaluating the rendered-to-image discrepancy.
The results show that the choice of representation strongly influences the optimization landscape.
3DGS~\cite{kerbl20233dgs} produces a rugged surface due to view-dependent overfitting, leading to unstable pose updates.
In contrast, our representation preserves global structure via anchor Gaussians while reducing local ambiguity with auxiliary Gaussians, resulting in a smoother landscape and more reliable convergence (see Fig.~\ref{fig:loss_landscape}).

\subsection{Adaptive Voxel Control}
\label{sec:adaptive_voxel_control}
To balance spatial resolution and computational efficiency, 
we select the voxel size $\varepsilon^*$ such that the number of voxels after downsampling matches a target value $V_{\text{target}}$, using a binary search:
\begin{equation}
\varepsilon^* = \argmin_{\varepsilon} \left|\, V(\varepsilon) - V_{\text{target}} \,\right|.
\end{equation}
The target voxel count $V_{\text{target}}$ is defined proportional to the LiDAR trajectory length, $D_{\text{traj}} = \sum_{t=1}^{T-1} \| \mathbf{T}^L_{t+1} - \mathbf{T}^L_t \|_2$, as $V_{\text{target}} = \beta\, D_{\text{traj}}$, where $\beta$ is a proportionality constant.
This formulation enables \textit{adaptive voxel control} (AVC), which dynamically regulates the number of anchor Gaussians according to the overall scene scale.
Importantly, AVC also influences calibration accuracy.
An excessively small $\varepsilon^*$ leads to over-densified anchors, which can hinder optimization stability.
Conversely, an excessively large $\varepsilon^*$ reduces geometric coverage and degrades calibration precision.
These effects are summarized in Tab.~\ref{tab:abl_model}.


\begin{table*}[t!]
    \centering
    \captionsetup{font=footnotesize}
    \caption{
        NVS results on \textsc{KITTI-360}~\cite{liao2022kitti360} and \textsc{Waymo}~\cite{sun2020waymo}.
        Metrics are averaged over 10 runs with color-coded rankings.
    }
    \resizebox{1.0\textwidth}{!}{
    \setlength{\tabcolsep}{2.5pt}
    \begin{tabular}{ll|*3{c}|*3{c}|*3{c}|*3{c}|*3{c}|*3{c}}
    \toprule[1.2pt]
    & \multirow{2}{*}{\small{Datasets}}
    & \multicolumn{3}{c|}{\small{Dataset Calib.}}
    & \multicolumn{3}{c|}{\small{CalibAnything~\cite{luo2023calibanything}}}
    & \multicolumn{3}{c|}{\small{INF~\cite{zhou2023inf}}}
    & \multicolumn{3}{c|}{\small{RobustCalib~\cite{zhou2025robust}}}
    & \multicolumn{3}{c|}{\small{3DGS-Calib$^*$~\cite{herau20243dgscalib}}}
    & \multicolumn{3}{c}{\small{Ours}} \\

    & & PSNR$\uparrow$ & SSIM$\uparrow$ & LPIPS$\downarrow$
    & PSNR$\uparrow$ & SSIM$\uparrow$ & LPIPS$\downarrow$
    & PSNR$\uparrow$ & SSIM$\uparrow$ & LPIPS$\downarrow$
    & PSNR$\uparrow$ & SSIM$\uparrow$ & LPIPS$\downarrow$
    & PSNR$\uparrow$ & SSIM$\uparrow$ & LPIPS$\downarrow$
    & PSNR$\uparrow$ & SSIM$\uparrow$ & LPIPS$\downarrow$ \\
    
    \midrule
    \multirow{6}{*}{\rotatebox{90}{\textsc{KITTI-360~~}}} 
    & \texttt{Straight} & \second{26.29} & \second{0.85} & \best{0.08} & {23.56} & {0.78} & {0.15} & {25.45} & \third{0.83} & \third{0.11} & {25.43} & {\third{0.83}} & {\second{0.10}} & \third{25.59} & \third{0.83} & \second{0.10} & \best{26.47} & \best{0.86} & \best{0.08} \\
    
    & \texttt{Small zigzag} & \second{27.13} & \second{0.89} & \best{0.07} & {23.96} & {0.83} & {0.12} & {24.21} & {0.83} & \third{0.13} & {26.04} & {0.87} & {\second{0.09}} & \third{26.48} & \third{0.88} & \second{0.09} & \best{27.38} & \best{0.90} & \best{0.07} \\
    
    & \texttt{Large zigzag} & \second{26.87} & \second{0.86} & \best{0.09} & {18.03} & {0.64} & {0.32} & \third{26.27} & \third{0.85} & \second{0.10} & {25.31} & {0.83} & {\third{0.13}} & {22.89} & {0.78} & {0.17} & \best{27.06} & \best{0.87} & \best{0.09} \\
    
    & \texttt{Small rotation} & \second{24.96} & \best{0.81} & \best{0.13} & {13.58} & {0.50} & {0.52} & {23.50} & \third{0.77} & \third{0.17} & {18.52} & {0.65} & {0.32} & \third{24.26} & \second{0.79} & \second{0.15} & \best{25.00} & \best{0.81} & \best{0.13} \\
    
    & \texttt{Large rotation} & \second{25.77} & \second{0.83} & \second{0.11} & {22.00} & {0.72} & {0.20} & {22.80} & {0.76} & {0.17} & {18.56} & {0.65} & {0.31} & \third{24.55} & \third{0.80} & \third{0.14} & \best{26.06} & \best{0.84} & \best{0.10} \\
    
    \cmidrule(){2-20}
    & Mean & \second{26.20} & \best{0.85} & \second{0.10} & {20.22} & {0.69} & {0.26} & {24.45} & \third{0.81} & {0.14} & {22.77} & {0.77} & {0.19} & \third{24.52} & \second{0.80} & \third{0.14} & \best{26.39} & \best{0.85} & \best{0.09} \\
    
    \midrule[0.8pt]
    \multirow{4}{*}{\rotatebox{90}{\textsc{Waymo~~}}} 
    & \texttt{Scene 81} & {27.81} & \second{0.88} & \third{0.13} & {25.66} & {{0.84}} & {0.18} & \third{28.08} & \second{0.88} & \second{0.12} & {27.11} & {\third{0.86}} & {0.14} & \second{28.64} & \best{0.89} & \best{0.11} & \best{29.06} & \best{0.89} & \best{0.11} \\
    
    & \texttt{Scene 226} & \third{23.78} & \third{0.77} & \second{0.18} & {17.59} & {0.61} & {0.40} & {22.10} & {0.75} & \third{0.22} & {\second{23.99}} & {\second{0.78}} & {\second{0.18}} & {23.70} & \third{0.77} & \second{0.18} & \best{24.97} & \best{0.80} & \best{0.15} \\
    
    & \texttt{Scene 362} & {25.99} & \third{0.87} & \third{0.11} & {17.87} & {0.72} & {0.33} & {20.89} & {0.78} & {0.24} & {\second{26.79}} & {\second{0.88}} & {\second{0.10}} & \third{26.65} & \second{0.88} & \best{0.09} & \best{27.08} & \best{0.89} & \best{0.09} \\
    
    \cmidrule(){2-20}
    & Mean & {25.86} & \third{0.84} & \third{0.14} & {20.37} & {0.72} & {0.31} & {23.69} & {0.80} & {0.19} & {\third{25.96}} & {\third{0.84}} & {\third{0.14}} & \second{26.33} & \second{0.85} & \second{0.13} & \best{27.04} & \best{0.86} & \best{0.11} \\

    \bottomrule[1.2pt]
    \end{tabular}}
    \label{tab:nvs}
    \vspace{-2mm}
\end{table*}
\begin{table}[t]
    \centering
    \captionsetup{font=footnotesize}
    \caption{
        NVS and calibration accuracy on \textsc{Fast-LIVO2}~\cite{zheng2024fastlivo2}.
        Results using the dataset calibration are included for reference.
    }
    \resizebox{1.0\columnwidth}{!}{
    \setlength{\tabcolsep}{1.0pt}
    \begin{tabular}{l|ccc|ccc|cc}
    \toprule[1.2pt]
    \multirow{2}{*}{\small{Scenes}} & \multicolumn{3}{c|}{\small{Dataset Calib.}} & \multicolumn{5}{c}{\small{Ours}} \\
     & {PSNR$\uparrow$} & {SSIM$\uparrow$} & {LPIPS$\downarrow$} & {PSNR$\uparrow$} & {SSIM$\uparrow$} & {LPIPS$\downarrow$} & {$\mathbf{R}$($^\circ$)$\downarrow$} & {$\mathbf{t}$(cm)$\downarrow$} \\
    
    \midrule
    \texttt{Building} & {22.13} & \textbf{0.755} & {0.172} & \textbf{22.16} & {0.754} & \textbf{0.171} & {0.56$\pm$0.01} & {14.5$\pm$0.37} \\
    
    \texttt{Landmark} & {23.81} & {0.656} & {0.223} & \textbf{23.99} & \textbf{0.660} & \textbf{0.220} & {0.15$\pm$0.01} & {8.71$\pm$0.08} \\
    
    \texttt{Sculpture} & {19.47} & {0.550} & \textbf{0.224} & \textbf{19.56} & \textbf{0.553} & \textbf{0.224} & {0.46$\pm$0.02} & {10.1$\pm$0.41} \\

    \midrule
    {Mean} & {21.81} & {0.654} & {0.207} & {\textbf{21.90}} & {\textbf{0.656}} & {\textbf{0.205}} & {0.39$\pm$0.18} & {11.1$\pm$2.51} \\
    
    \bottomrule[1.2pt]
    \end{tabular}}    
    \label{tab:fastlivo2}
    \vspace{-3mm}
\end{table}

\subsection{Loss Function}
We define the total loss as a combination of photometric supervision and a regularization term:
\begin{equation}
\mathcal{L}_{\text{total}} = \lambda_{\text{photo}}\mathcal{L}_{\text{photo}} + \lambda_{\text{scale}}\mathcal{L}_{\text{scale}}.
\end{equation}

The photometric loss $\mathcal{L}_{\text{photo}}$ follows~\cite{kerbl20233dgs}, combining \textit{L1} and \textit{D-SSIM} terms for image reconstruction.
The regularization term $\mathcal{L}_{\text{scale}}$ penalizes degenerate, highly anisotropic Gaussians by constraining their aspect ratios.
It is applied only to Gaussians that pass the view frustum filter.
Let $\mathcal{V}$ be the set of valid Gaussians, with each $\mathbf{s}_i \in \mathbb{R}^3$ representing scale along each axis. The loss is defined as:
\begin{equation}
    \mathcal{L}_{\text{scale}} = 
    \frac{1}{|\mathcal{V}|} 
    \sum_{i \in \mathcal{V}} 
    \max \left(\frac{\max(\mathbf{s}_i)}{\min(\mathbf{s}_i)} - \sigma, \; 0 \right),
\label{eq:12}
\end{equation}
where $\sigma$ is a predefined threshold (see Sec.~\ref{sec:implementation_details}).
If $|\mathcal{V}| = 0$, the loss is set to zero.
This term stabilizes training by softly constraining Gaussian shapes while allowing adaptation to local geometry.

\section{Experimental evaluation}
\label{sec:experiments}

\subsection{Dataset}
\label{sec:dataset}

We evaluate our method on two public autonomous driving datasets, \textsc{KITTI-360}~\cite{liao2022kitti360} and \textsc{Waymo}~\cite{sun2020waymo}, as well as one handheld dataset, \textsc{Fast-LIVO2}~\cite{zheng2024fastlivo2}.
\textsc{KITTI-360} includes a spinning LiDAR, two front-facing perspective cameras, and two side-mounted fisheye cameras.
%
Following prior work~\cite{herau20243dgscalib}, we select three representative scenes based on motion patterns: \texttt{Straight}, \texttt{Small zigzag}, and \texttt{Small rotation}.
To further examine more diverse motion patterns, we additionally include \texttt{Large zigzag} and \texttt{Large rotation}.
\textsc{Waymo} features a top-mounted spinning LiDAR and five perspective cameras covering the front and sides.
\textsc{Fast-LIVO2} consists of a solid-state LiDAR paired with a single perspective camera, with both sensors facing forward.
For evaluation, we select three test scenes from each dataset that contain fewer dynamic objects: \textsc{Waymo} (\texttt{81}, \texttt{226}, \texttt{362}) and \textsc{Fast-LIVO2} (\texttt{Building}, \texttt{Landmark}, \texttt{Sculpture}).
For each scene, we select approximately 80 sequential images per camera, using every second frame as a training view.


\subsection{Experimental Setup}

\subsubsection{Initialization}
%
To evaluate our method on three datasets, we follow the \textit{from-LiDAR initialization} protocol~\cite{yang2024unical}, which provides coarse camera poses derived from LiDAR odometry and approximate camera rotations.
For instance, in \textsc{KITTI-360}, the four cameras are roughly aligned with yaw angles of 0\textdegree{} (front), 90\textdegree{} (left), and –90\textdegree{} (right).
This initialization is challenging, as translation errors can reach up to $1.2$m, substantially degrading calibration accuracy.
As a result, existing targetless calibration methods often fail to converge under this setting, as described in Sec.~\ref{sec:eval_calibration_accuracy}.

\subsubsection{Baselines}

We compare our method against four baselines, along with the dataset-provided calibration, resulting in a total of five comparison methods.
Baselines are selected based on two criteria: (i) the availability of publicly released implementations to ensure consistent and reproducible evaluation, and (ii) relevance to targetless LiDAR-camera calibration.
We adopt these criteria because several closely related approaches~\cite{herau2023moisst, herau20243dgscalib, yang2024unical} have not released their source code, making direct and reproducible comparison difficult.
As an exception, we include 3DGS-Calib~\cite{herau20243dgscalib}, which is the most closely related method to our work.
Since its official implementation is unavailable, we re-implement it following the paper and include it as our primary baseline, denoted with an asterisk ($^*$).
Implementation details are provided in Sec.~\ref{sec:implementation_details_baselines}.
%
Finally, we include the \textit{dataset calibration} officially released with the dataset as a reference for comparison.
%


\begin{figure*}[t]
    \centering
    \captionsetup{font=footnotesize}
    \includegraphics[width=0.99\linewidth]{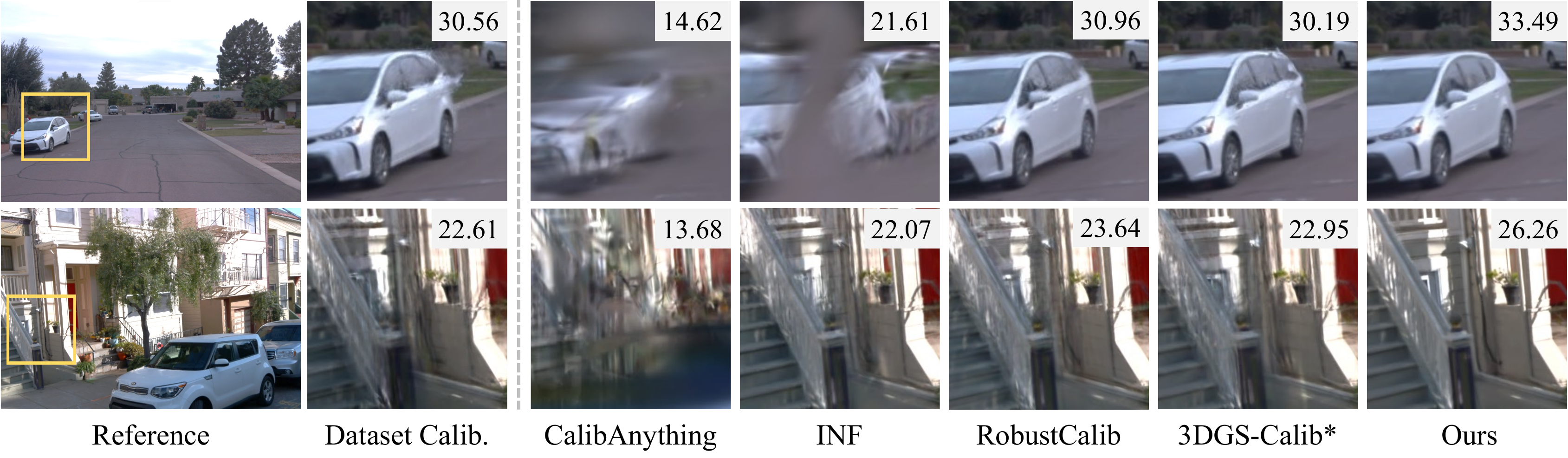}
    \caption{
        Qualitative comparison of Novel View Synthesis results on \textsc{Waymo}.
        Key improvements are highlighted with yellow boxes, and cropped patches show zoomed-in regions for clarity.
        The PSNR of each rendered image is shown in the top-right corner.
    }
    \label{fig:waymo_render}
\end{figure*}
\begin{figure*}[t]
    \centering
    \captionsetup{font=footnotesize}
    \includegraphics[width=0.99\linewidth]{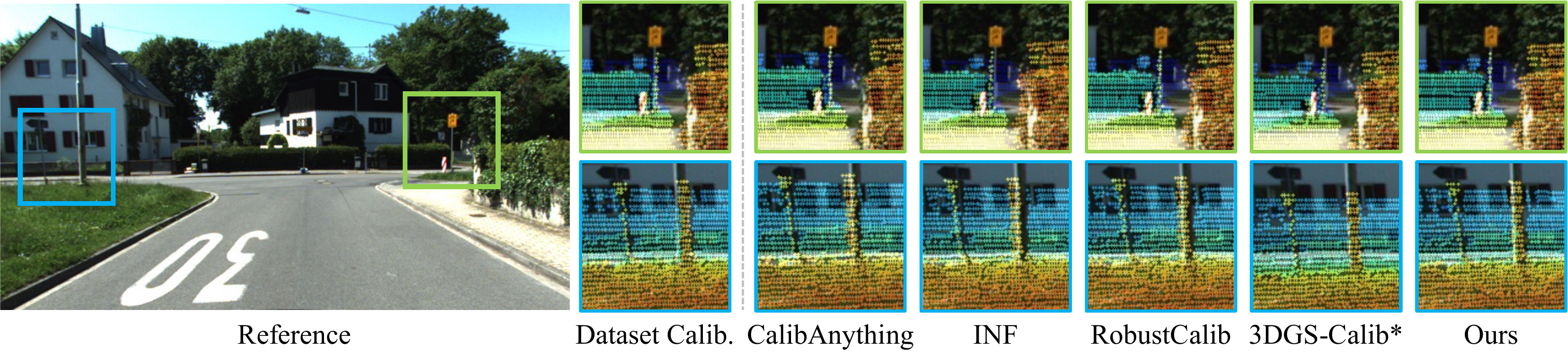}
    \caption{
        Qualitative evaluation of LiDAR-camera alignment on \textsc{KITTI-360}.
        LiDAR points are projected onto images using the calibration results estimated by each baseline method.
        Point colors indicate 3D distances from the LiDAR, ranging from red (near) to blue (far).
    }
    \label{fig:kitti360_prj}
    \vspace{-2mm}
\end{figure*}

\subsection{Implementation Details}
\label{sec:implementation_details}

\subsubsection{\ours{}}
\label{sec:implementation_details_ours}
We use a two-layer MLP with ReLU activation and 32 hidden units to regress Gaussian attributes.
The parameters are set to $K{=}5$, $\beta{=}5000$, and $\sigma{=}10$ for the number of auxiliary Gaussians, the proportionality factor of $N_{\text{target}}$, and the scale regularization threshold.
Training is conducted for 30K iterations using AdamW, with a weight decay of $10^{-2}$ for the first 15K iterations.
Each camera has a separate optimizer, with learning rates of $2{\times}10^{-3}$ for rotation and $5{\times}10^{-3}$ for translation, using a cosine annealing scheduler that decays the learning rate to $0.1{\times}$ the initial value.
The loss combines \textit{D-SSIM} and scale terms, weighted by $\lambda_{\text{D-SSIM}}{=}0.2$ and $\lambda_{\text{scale}}{=}1.0$.
With this configuration, the method remains below 8GB of VRAM and uses the \emph{same hyperparameters} across all datasets.
All experiments are averaged over 10 random seeds and are conducted on a single RTX 4090 GPU.

\subsubsection{Baselines}
\label{sec:implementation_details_baselines}

For 3DGS-Calib$^*$~\cite{herau20243dgscalib}, our primary baseline, we follow the original implementation with only minor adjustments.
We use the standard hash-grid encoding with scene contraction and lightweight MLP heads with a hidden dimension 64.
Based on direct communication with the original authors, we adopt pose learning rates of $1.5 \times 10^{-2}$ for translation and $1.5 \times 10^{-3}$ for rotation.
Due to memory requirements, this baseline is evaluated on an RTX 5090 GPU.
For baselines that assume a single camera setup~\cite{luo2023calibanything, zhou2025robust}, we run the method independently for each camera and aggregate the results across cameras for evaluation.

\subsection{Evaluation of Calibration Accuracy}
\label{sec:eval_calibration_accuracy}
Calibration accuracy is evaluated by comparing the estimated extrinsics with the \textit{dataset calibration}.
We report the success rate (SR, \%), rotation error ($^\circ$), and translation error (cm), where a calibration is considered successful if the rotation and translation errors are within $1^\circ$ and 20\,cm, respectively~\cite{zhou2025robust}.
Tab.~\ref{tab:kitti360_acc} summarizes the results on the \textsc{KITTI-360}~\cite{liao2022kitti360} dataset.
Our method achieves state-of-the-art performance across all evaluated scenes, attaining a 100\% SR.
In contrast, baseline methods succeed only in limited motion scenarios, with the strongest baseline, INF~\cite{zhou2023inf}, reaching at most 74.5\% SR.

\textsc{KITTI-360} includes both front-view perspective and side-mounted fisheye cameras, requiring consistent calibration across wide viewing angles and severe distortions.
Nevertheless, our method maintains high accuracy across all camera types, improving rotation and translation accuracy by 68.3\% and 62.5\%, respectively.
Moreover, the average rotation error of our method remains around $0.1^\circ$, effectively reducing large projection deviations caused by small angular misalignments.
Fig.~\ref{fig:kitti360_prj} qualitatively visualizes the alignment quality, showing that our method preserves accurate alignment even for distant objects.
This robustness stems from jointly optimizing camera poses within a shared scene representation that enforces structural consistency across views.
In contrast, baseline methods~\cite{luo2023calibanything, zhou2023inf, herau20243dgscalib, zhou2025robust} often exhibit unstable convergence depending on hyperparameter choices and scene complexity.
Tab.~\ref{tab:time_comparison} further reports the training time comparison.
Our method is the second fastest among all approaches, with an average training time of \mbox{$\sim$}0.18 hours.
Additional calibration results on the \textsc{Fast-LIVO2}~\cite{zheng2024fastlivo2} dataset are provided in Tab.~\ref{tab:fastlivo2}.
Finally, \textsc{Waymo}~\cite{sun2020waymo} provides less reliable calibration and pose information, as reported in recent work~\cite{herau2025pose}.
We therefore evaluate calibration quality on this dataset using Novel View Synthesis (NVS) performance as a complementary indicator to assess extrinsic alignment.



\subsection{Evaluation of Novel View Synthesis}
\label{sec:eval_nvs}


Recent studies~\cite{brachmann2024scene} have shown that accurate camera poses are critical for achieving high-quality NVS.
Accordingly, in addition to explicit pose accuracy metrics, we evaluate NVS performance to assess the impact of calibration accuracy on rendering consistency.
For a fair comparison, we employ a unified 3DGS-based NVS pipeline~\cite{kerbl20233dgs}, using LiDAR points for initialization.
All methods are evaluated within this identical pipeline, ensuring that NVS differences arise solely from the quality of the estimated LiDAR-to-camera extrinsics.
As shown in Tab.~\ref{tab:nvs}, our method achieves the highest rendering quality across all scenes on both \textsc{KITTI-360} and \textsc{Waymo}.
On the \textsc{Waymo} dataset, where the provided calibration is less reliable~\cite{herau2025pose}, baseline methods such as RobustCalib~\cite{zhou2025robust} and 3DGS-Calib~\cite{herau20243dgscalib} already outperform the dataset calibration in NVS quality.
In contrast, on \textsc{KITTI-360}, which provides highly accurate ground-truth data, our method is the only approach that consistently surpasses the dataset calibration in NVS performance, indicating improved extrinsic estimation beyond the provided reference.
Fig.~\ref{fig:waymo_render} qualitatively illustrates this advantage, showing that our calibration produces sharper reconstructions and more consistent rendering compared to both baseline methods and the dataset calibration.
To further assess the sensor generalization capability of our approach, we additionally evaluate it on a solid-state LiDAR setup, which has not been explored by prior calibration methods.
As reported in Tab.~\ref{tab:fastlivo2}, consistent with previous results, our method slightly outperforms the dataset calibration.




\subsection{Ablation Study}

\subsubsection{Robustness to Noisy Initialization}
\label{sec:robustness}
\begin{figure}[t]
    \centering
    \captionsetup{font=footnotesize}
    \includegraphics[width=0.99\linewidth]{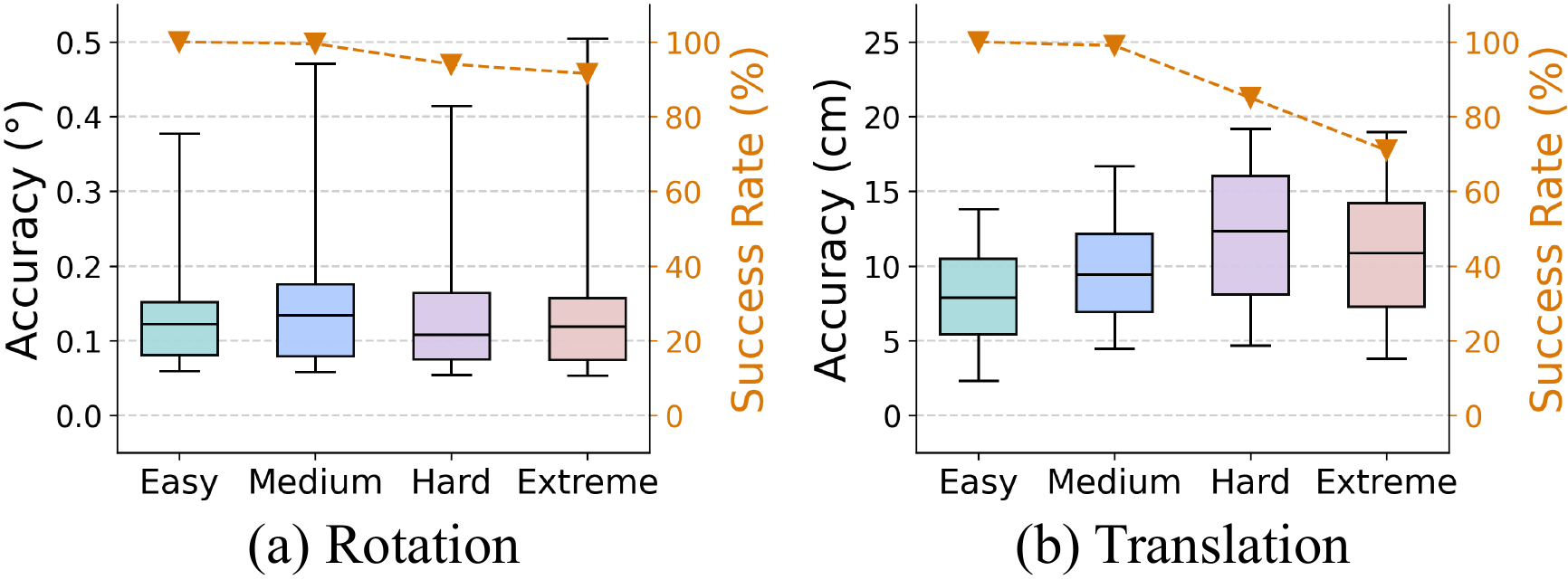}
    \caption{
        Robustness to initialization perturbations.
        We evaluate four difficulty levels: \texttt{Easy}, \texttt{Medium}, \texttt{Hard}, and \texttt{Extreme}.
        For each level, we report the success rate (right axis) and the calibration error (left axis) over successful runs for (a) rotation and (b) translation.
    }
    \label{fig:kitti360_perturb}
    \vspace{-5mm}
\end{figure}

We analyze the robustness of our method to noisy initialization by progressively increasing the magnitude of pose perturbations.
As shown in Fig.~\ref{fig:kitti360_perturb}, we evaluate on four different difficulty levels: \texttt{Easy}, \texttt{Medium}, \texttt{Hard}, and \texttt{Extreme}, where the initial extrinsics are perturbed by up to
($5^\circ$, 0.5\,m), ($10^\circ$, 1.0\,m), ($15^\circ$, 1.5\,m), and ($20^\circ$, 2.0\,m), respectively.
For each level, we report the success rate and the calibration error over successful runs for both rotation and translation.
As the perturbation magnitude increases, the success rate gradually decreases, with the most noticeable drop under the \texttt{Extreme} setting.

However, across all difficulty levels, the median rotation and translation errors of successful runs remain consistently low, indicating stable convergence once optimization succeeds.
In practical scenarios, initial calibration errors typically fall within the \texttt{Easy} or \texttt{Medium} ranges.
Under these conditions, our method achieves near-perfect success rates with consistently low calibration errors.
These results show that our approach is well suited for real-world targetless calibration, where initial poses are often imprecise but rarely subject to large perturbations.


\begin{table}[t!]
    \centering
    \captionsetup{font=footnotesize}
    \caption{
        Ablation study on model components and voxel density.
        Calibration accuracy and training time are reported for varying $\epsilon$ and $K$, with AVC selecting voxel sizes automatically.
    }
    \resizebox{1.0\columnwidth}{!}{
    \begin{tabular}{cccc|cc|c}
    \toprule[1.2pt]
    \multicolumn{4}{c|}{{Methods}} & \multicolumn{2}{c|}{{Accuracy}} & \multirow{2}{*}{Time(s)$\downarrow$} \\
    R-O & $\mathcal{L}_\text{scale}$ & AVC & $K$ / $\epsilon$ & $\mathbf{R}$($^\circ$)$\downarrow$ & $\mathbf{t}$(cm)$\downarrow$ & \\
    
    \midrule
    & \checkmark & \checkmark & {5 / -}  & {1.94} & {64.7} & {647} \\
    \checkmark & & \checkmark & {5 / -}  & {0.24} & {11.2} & {624} \\
    
    {\checkmark} & {\checkmark} & & {5} / {0.5} & {0.15}  & {16.4} & {\textbf{591}} \\
    {\checkmark} & {\checkmark} & & {5} / {0.3} & {\underline{0.13}}  & {\underline{9.13}} & {\underline{612}} \\
    \checkmark & \checkmark & & {5 / 0.1} & {0.14}  & {9.74} & {673} \\

    {\checkmark} & {\checkmark} & {\checkmark} & {20 / -}  & \textbf{{0.12}} & {9.99} & {729} \\
    {\checkmark} & {\checkmark} & {\checkmark} & {10 / -}  & \underline{{0.13}} & {9.71} & {680} \\

    \midrule
    \checkmark & \checkmark & \checkmark & {5 / -}   & \underline{{0.13}} & \textbf{{8.86}} & {625} \\
    
    \bottomrule[1.2pt]
    \end{tabular}}    
    \label{tab:abl_model}
    \vspace{-3mm}
\end{table}

\begin{figure}[t]
    \centering
    \captionsetup{font=footnotesize}
    \includegraphics[width=1.0\linewidth]{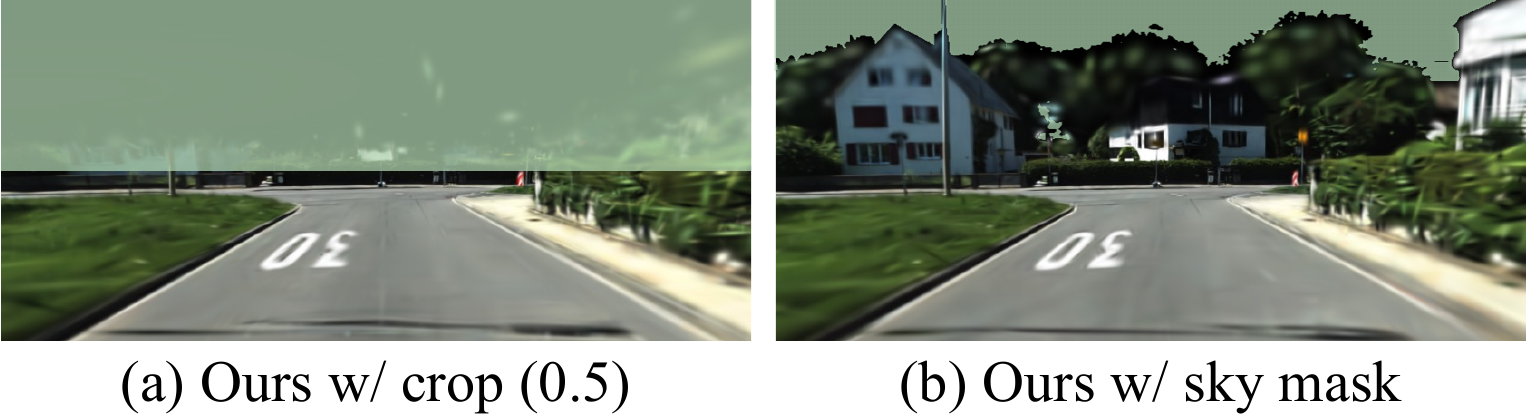}    
    \caption{
        Rendering results of our method with (a) image cropping (0.5) and (b) sky masking.
        The mask is overlaid for visualization.
    }
    \label{fig:mask_visualize}
    \vspace{-5mm}
\end{figure}
\begin{table}[t]
    \centering
    \captionsetup{font=footnotesize}
    \caption{
        Effect of masking strategies on calibration and NVS metrics.
        Calibration accuracy and NVS performance under different masking settings.
    }
    \resizebox{1.0\columnwidth}{!}{
    \setlength{\tabcolsep}{4pt}
    \begin{tabular}{l|cc|ccc}
    \toprule[1.2pt]
    \multirow{2}{*}{{Models}} & \multicolumn{2}{c|}{{Accuracy}} & \multicolumn{3}{c}{{Novel View Synthesis}} \\
    & $\mathbf{R}$($^\circ$)$\downarrow$ & $\mathbf{t}$(cm)$\downarrow$ & PSNR$\uparrow$ & SSIM$\uparrow$ & LPIPS$\downarrow$ \\
    
    \midrule    
    Ours w/ crop (0.7) & 1.64 & 72.9 & 22.16 & 0.751 & 0.196 \\    
    Ours w/ crop (0.5) & 0.59 & 40.9 & 25.87 & 0.841 & 0.106 \\    
    Ours w/ crop (0.3) & 0.14 & 22.1 & \underline{26.27} & 0.851 & 0.097 \\    
    Ours w/ sky mask & \textbf{0.12} & \underline{17.5} & 26.26 & \underline{0.852} & \underline{0.096} \\
    \textbf{Ours full} & \underline{0.13} & \textbf{8.86} & \textbf{26.39} & \textbf{0.854} & \textbf{0.094} \\
    
    \bottomrule[1.2pt]
    \end{tabular}}
    \label{tab:effect_mask}
    \vspace{-4mm}
\end{table}

\begin{table}[t!]
    \centering
    \captionsetup{font=footnotesize}
    \caption{
        Ablation study of extrinsic parameter generalization on \textsc{KITTI-360}~\cite{liao2022kitti360}.   
        Calibration optimized on \boxcolorref{\texttt{Scene A}} is applied to \texttt{Scene B} and \texttt{Scene C}, and evaluated using NVS metrics.
    }
    \resizebox{1.0\columnwidth}{!}{
    \setlength{\tabcolsep}{4pt}
    \begin{tabular}{l|ccc|ccc}
    \toprule[1.2pt]
    \multirow{2}{*}{\small{Scenes}} & \multicolumn{3}{c|}{\small{Dataset Calib.}} & \multicolumn{3}{c}{\small{Ours}} \\
     
     & {PSNR$\uparrow$} & {SSIM$\uparrow$} & {LPIPS$\downarrow$} & {PSNR$\uparrow$} & {SSIM$\uparrow$} & {LPIPS$\downarrow$} \\
    
    \midrule
    \bxref{\texttt{Seq. 9A}} & {26.87} & {0.864} & {0.092} & {\textbf{27.06}} & {\textbf{0.867}} & {\textbf{0.091}} \\
    \texttt{Seq. 9B} & 26.21 & 0.866 & \textbf{0.089} & \textbf{26.37} & \textbf{0.868} & \textbf{0.089} \\
    \texttt{Seq. 9C} & 24.01 & 0.829 & 0.120 & \textbf{24.08} & \textbf{0.830} & \textbf{0.119} \\
    
    \midrule
    \bxref{\texttt{Seq. 10A}} & {24.96} & {0.810} & {0.130} & {\textbf{25.00}} & {\textbf{0.813}} & {\textbf{0.129}} \\
    \texttt{Seq. 10B} & 24.04 & 0.814 & {0.124} & \textbf{24.25} & \textbf{0.817} & \textbf{0.123} \\
    \texttt{Seq. 10C} & 23.74 & {0.832} & {0.100} & \textbf{23.95} & \textbf{0.836} & \textbf{0.098} \\
    
    \bottomrule[1.2pt]
    \end{tabular}}
    \label{tab:abl_general}
    \vspace{-3mm}
\end{table}

\subsubsection{Model Components and Voxel Density}
\label{sec:model_comp_voxel}
Tab.~\ref{tab:abl_model} analyzes the impact of model components and voxel density on calibration accuracy and training time.
Removing rig optimization (R-O) results in large rotation and translation errors, confirming that joint optimization of camera extrinsics is essential.
Scale regularization further improves optimization stability and consistently reduces calibration error.
Increasing the number of auxiliary Gaussians $K$ to 2$\times$ or 4$\times$ the default value ($K{=}5$) reduces the relative contribution of anchor Gaussians, leading to degraded translation accuracy.
With manually fixed voxel sizes $\epsilon$, calibration performance becomes sensitive to hyperparameter choices, and overly fine voxelization increases training time with limited accuracy gains.
In contrast, adaptive voxel control (AVC presented in Sec.~\ref{sec:adaptive_voxel_control}) achieves comparable or better accuracy without requiring manual tuning of $\epsilon$.
Overall, our method exhibits limited sensitivity to voxel density, with only minor performance degradation observed under extreme settings.

\subsubsection{Effect of Masking LiDAR-Unobserved Areas}
\label{sec:effect_mask}
Prior 3DGS-based targetless calibration methods~\cite{herau20243dgscalib, zhou2025robust} restrict photometric supervision to LiDAR-observed regions in order to maintain geometric consistency, thereby discarding image regions that are not directly supported by LiDAR measurements (Fig.~\ref{fig:scene_rep_comp}).
To analyze how this design choice affects optimization behavior and calibration accuracy, we conduct an ablation study that explicitly masks LiDAR-unobserved image regions.
Tab.~\ref{tab:effect_mask} summarizes calibration accuracy and NVS performance under different masking strategies.
Both image cropping and sky masking remove photometric supervision from LiDAR-unobserved regions, reducing the amount of available image information and degrading pose optimization.
In particular, calibration performance consistently deteriorates as the cropping ratio increases, and even a relatively mild cropping ratio (0.3) leads to a noticeable performance drop compared to the full model.
Similarly, sky masking~\cite{xie2021segformer} also underperforms the full model, as it restricts a comparable amount of photometric information to mild image cropping.
In contrast, the full model achieves the best calibration accuracy and NVS quality by retaining photometric supervision over the entire image without explicit masking.
Fig.~\ref{fig:mask_visualize} illustrates the masking strategies used in this ablation study, while Fig.~\ref{fig:scene_rep_comp}(c) shows that our method successfully reconstructs upper image regions without masking.

\subsubsection{Extrinsic Generalization}
\label{sec:extrinsic_generalization}
To evaluate extrinsic generalization, we apply the extrinsics optimized on \boxcolorref{\texttt{Scene A}} to novel \texttt{Scene B} and \texttt{Scene C} without recalibration.  
A key indicator of calibration quality is whether the estimated extrinsics remain effective when transferred across sequences captured by the same vehicle.  
Here, \boxcolorref{\texttt{Seq.~9A}} (\texttt{Large zigzag}) and \boxcolorref{\texttt{Seq.~10A}} (\texttt{Small rotation}) serve as source scenes, while \texttt{9B/C} and \texttt{10B/C} are their corresponding targets within the same \textsc{KITTI-360} sequence.
For clarity, source scenes are indicated in \boxcolorref{\texttt{gray}}.
As shown in Tab.~\ref{tab:abl_general}, our method achieves consistently higher NVS metrics on the target scenes than the default dataset calibration, demonstrating strong cross-scene calibration robustness.

\section{Conclusion}
\label{sec:conclusion}

In this paper, we introduced \ours{}, a targetless LiDAR–camera calibration that leverages a neural scene representation without scene-specific hyperparameters.
Our approach jointly optimizes the scene representation and sensor poses using neural Gaussians, which mitigate viewpoint overfitting, avoid poor local minima, and improve overall optimization stability.
Experiments on two public driving datasets and one handheld dataset demonstrate that \ours{} achieves superior pose accuracy, rendering quality, and generalization compared to existing methods.
Similar to prior targetless calibration approaches, our method assumes synchronized sensors, and relies on reasonably accurate LiDAR odometry for initialization.
Extending the framework to handle temporal misalignment, dynamic environments, and joint multi-LiDAR calibration remains an important direction for future work.


\addtolength{\textheight}{-0cm}   





{\small
    \bibliographystyle{IEEEtran}
    \bibliography{reference}

@String(PAMI = {IEEE Trans. Pattern Anal. Mach. Intell.})

@String(IJRR = {Int. J. Robot. Res.})

@String(ECCV = {Eur. Conf. Comput. Vis.})

@String(CVPR = {IEEE Conf. Comput. Vis. Pattern Recog.})

@String(NIPS = {Adv. Neural Inform. Process. Syst.})

@String(TOG  = {ACM Trans. Graph.})

@String(TVCG = {IEEE Trans. Vis. Comput. Graph.})

@String(ICLR = {Int. Conf. Learn. Represent.})

@String(CVPRW= {IEEE Conf. Comput. Vis. Pattern Recog. Worksh.})

@String(IVS  = {IEEE Intell. Vehi. Symp.})

@String(IROS = {IEEE/RSJ Int. Conf. Intell. Robot. Syst.})

@String(ICRA = {IEEE Int. Conf. Robot. Automat.})

@String(RSS  = {Robot. Sci. Syst.})

@String(RAL  = {IEEE Robot. Automat. Lett.})

@String(TRO  = {IEEE Trans. Robot.})

@inproceedings{mildenhall2020nerf,
  title={{NeRF: Representing scenes as neural radiance fields for view synthesis}},
  author={Mildenhall, Ben and Srinivasan, Pratul P and Tancik, Matthew and Barron, Jonathan T and Ramamoorthi, Ravi and Ng, Ren},
  booktitle=ECCV,
  pages={405--421},
  year={2020},
}

@inproceedings{zhou2023inf,
  title={{INF: Implicit neural fusion for lidar and camera}},
  author={Zhou, Shuyi and Xie, Shuxiang and Ishikawa, Ryoichi and Sakurada, Ken and Onishi, Masaki and Oishi, Takeshi},
  booktitle=IROS,
  pages={10918--10925},
  year={2023}
}

@inproceedings{herau2023moisst,
  title={{MOISST: Multimodal Optimization of Implicit Scene for SpatioTemporal Calibration}},
  author={Herau, Quentin and Piasco, Nathan and Bennehar, Moussab and Rold{\~a}o, Luis and Tsishkou, Dzmitry and Migniot, Cyrille and Vasseur, Pascal and Demonceaux, C{\'e}dric},
  booktitle=IROS,
  pages={1810--1817},
  year={2023}
}

@inproceedings{yang2024unical,
  title={{UniCal: Unified Neural Sensor Calibration}},
  author={Yang, Ze and Chen, George and Zhang, Haowei and Ta, Kevin and B{\^a}rsan, Ioan Andrei and Murphy, Daniel and Manivasagam, Sivabalan and Urtasun, Raquel},
  booktitle=ECCV,
  pages={327--345},
  year={2024},
}

@article{zwicker2002ewa,
  title={{EWA splatting}},
  author={Zwicker, Matthias and Pfister, Hanspeter and Van Baar, Jeroen and Gross, Markus},
  journal=TVCG,
  volume={8},
  number={3},
  pages={223--238},
  year={2002},
}

@article{kerbl20233dgs,
  title={{3D gaussian splatting for real-time radiance field rendering}},
  author={Kerbl, Bernhard and Kopanas, Georgios and Leimk{\"u}hler, Thomas and Drettakis, George},
  journal=TOG,
  volume={42},
  number={4},
  pages={139--1},
  year={2023}
}

@inproceedings{herau20243dgscalib,
  title={{3DGS-Calib: 3D Gaussian Splatting for Multimodal SpatioTemporal Calibration}}, 
  author={Herau, Quentin and Bennehar, Moussab and Moreau, Arthur and Piasco, Nathan and Roldão, Luis and Tsishkou, Dzmitry and Migniot, Cyrille and Vasseur, Pascal and Demonceaux, Cédric},
  booktitle=IROS, 
  year={2024},
  pages={8315-8321},
}

@inproceedings{matsuki2024monogs,
  title={{Gaussian splatting slam}},
  author={Matsuki, Hidenobu and Murai, Riku and Kelly, Paul HJ and Davison, Andrew J},
  booktitle=CVPR,
  pages={18039--18048},
  year={2024}
}

@inproceedings{lu2024scaffold,
  title={{Scaffold-gs: Structured 3d gaussians for view-adaptive rendering}},
  author={Lu, Tao and Yu, Mulin and Xu, Linning and Xiangli, Yuanbo and Wang, Limin and Lin, Dahua and Dai, Bo},
  booktitle=CVPR,
  pages={20654--20664},
  year={2024},
}

@inproceedings{zhao2024tclc,
  title = {{TCLC-GS: Tightly Coupled LiDAR-Camera Gaussian Splatting for Autonomous Driving}},
  author = {Zhao, Cheng and Sun, Su and Wang, Ruoyu and Guo, Yuliang and Wan, Jun-Jun and Huang, Zhou and Huang, Xinyu and Chen, Yingjie Victor and Ren, Liu},
  booktitle = ECCV,
  year = {2024},
  pages = {91–106},
  numpages = {16},
}

@inproceedings{chen2025omnire,
  title={{OmniRe: Omni Urban Scene Reconstruction}},
  author={Ziyu Chen and Jiawei Yang and Jiahui Huang and Riccardo de Lutio and Janick Martinez Esturo and Boris Ivanovic and Or Litany and Zan Gojcic and Sanja Fidler and Marco Pavone and Li Song and Yue Wang},
  booktitle=ICLR,
  year={2025},
}

@article{zhou2025robust,
  title={{Robust LiDAR-Camera Calibration With 2D Gaussian Splatting}},
  author={Zhou, Shuyi and Xie, Shuxiang and Ishikawa, Ryoichi and Oishi, Takeshi},
  journal=RAL,
  year={2025},
}

@inproceedings{zhang2004extrinsic,
  title={{Extrinsic calibration of a camera and laser range finder}},
  author={Zhang, Qilong and Pless, Robert},
  booktitle=IROS,
  volume={3},
  pages={2301--2306},
  year={2004},
}

@inproceedings{scaramuzza2007extrinsic,
  title={{Extrinsic self calibration of a camera and a 3D laser range finder from natural scenes}},
  author={Scaramuzza, Davide and Harati, Ahad and Siegwart, Roland},
  booktitle=IROS,
  pages={4164--4169},
  year={2007},
}

@inproceedings{geiger2012automatic,
  title={{Automatic camera and range sensor calibration using a single shot}},
  author={Geiger, Andreas and Moosmann, Frank and Car, {\"O}mer and Schuster, Bernhard},
  booktitle=ICRA,
  pages={3936--3943},
  year={2012},
}

@article{mirzaei20123d,
  title={{3D LiDAR--camera intrinsic and extrinsic calibration: Identifiability and analytical least-squares-based initialization}},
  author={Mirzaei, Faraz M and Kottas, Dimitrios G and Roumeliotis, Stergios I},
  journal=IJRR,
  volume={31},
  number={4},
  pages={452--467},
  year={2012},
}

@inproceedings{levinson2013automatic,
  title={{Automatic online calibration of cameras and lasers}},
  author={Levinson, Jesse and Thrun, Sebastian},
  booktitle=RSS,
  volume={2},
  number={7},
  year={2013},
}

@article{munoz2020targetless,
  title={{Targetless camera-lidar calibration in unstructured environments}},
  author={Mu{\~n}oz-Ba{\~n}{\'o}n, Miguel {\'A}ngel and Candelas, Francisco A and Torres, Fernando},
  journal={IEEE Access},
  volume={8},
  pages={143692--143705},
  year={2020},
}

@inproceedings{schneider2017regnet,
  title={{RegNet: Multimodal sensor registration using deep neural networks}},
  author={Schneider, Nick and Piewak, Florian and Stiller, Christoph and Franke, Uwe},
  booktitle=IVS,
  pages={1803--1810},
  year={2017},
}

@inproceedings{lv2021lccnet,
  title={{LCCNet: LiDAR and camera self-calibration using cost volume network}},
  author={Lv, Xudong and Wang, Boya and Dou, Ziwen and Ye, Dong and Wang, Shuo},
  booktitle=CVPRW,
  pages={2888-2895},
  year={2021},
}

@inproceedings{luo2023calibanything,
  title={{Zero-training LiDAR-Camera Extrinsic Calibration Method Using Segment Anything Model}}, 
  author={Luo, Zhaotong and Yan, Guohang and Cai, Xinyu and Shi, Botian},
  booktitle=ICRA, 
  pages={14472-14478},
  year={2024},
}

@inproceedings{zhu2020online,
  title={{Online camera-lidar calibration with sensor semantic information}},
  author={Zhu, Yufeng and Li, Chenghui and Zhang, Yubo},
  booktitle=ICRA,
  pages={4970--4976},
  year={2020},
}

@inproceedings{zhang2021line,
  title={{Line-based automatic extrinsic calibration of LiDAR and camera}},
  author={Zhang, Xinyu and Zhu, Shifan and Guo, Shichun and Li, Jun and Liu, Huaping},
  booktitle=ICRA,
  pages={9347--9353},
  year={2021},
}

@article{rotter2022automatic,
  title={{Automatic calibration of a lidar--camera system based on instance segmentation}},
  author={Rotter, Pawel and Klemiato, Maciej and Skruch, Pawel},
  journal={Remote Sensing},
  volume={14},
  number={11},
  pages={2531},
  year={2022},
}

@inproceedings{levoy1996light,
  title = {{Light field rendering}},
  author = {Levoy, Marc and Hanrahan, Pat},
  year = {1996},
  publisher = {Association for Computing Machinery},
  booktitle = {Proceedings of the 23rd Annual Conference on Computer Graphics and Interactive Techniques},
  pages = {31–42},
  numpages = {12},
  series = {SIGGRAPH '96}
}

@inproceedings{gortler1996lumigraph,
  title = {{The lumigraph}},
  author = {Gortler, Steven J. and Grzeszczuk, Radek and Szeliski, Richard and Cohen, Michael F.},
  year = {1996},
  publisher = {Association for Computing Machinery},
  booktitle = {Proceedings of the 23rd Annual Conference on Computer Graphics and Interactive Techniques},
  pages = {43–54},
  numpages = {12},
  series = {SIGGRAPH '96}
}

@inproceedings{sun2020waymo,
  title={{Scalability in perception for autonomous driving: Waymo open dataset}},
  author={Sun, Pei and Kretzschmar, Henrik and Dotiwalla, Xerxes and Chouard, Aurelien and Patnaik, Vijaysai and Tsui, Paul and Guo, James and Zhou, Yin and Chai, Yuning and Caine, Benjamin and others},
  booktitle=CVPR,
  pages={2446--2454},
  year={2020}
}

@article{liao2022kitti360,
  title={{Kitti-360: A novel dataset and benchmarks for urban scene understanding in 2d and 3d}},
  author={Liao, Yiyi and Xie, Jun and Geiger, Andreas},
  journal=PAMI,
  volume={45},
  number={3},
  pages={3292--3310},
  year={2022},
}

@article{vizzo2023kiss,
  title={{Kiss-icp: In defense of point-to-point icp--simple, accurate, and robust registration if done the right way}},
  author={Vizzo, Ignacio and Guadagnino, Tiziano and Mersch, Benedikt and Wiesmann, Louis and Behley, Jens and Stachniss, Cyrill},
  journal=RAL,
  volume={8},
  number={2},
  pages={1029--1036},
  year={2023},
}

@article{chen2024ig,
  title={{ig-lio: An incremental gicp-based tightly-coupled lidar-inertial odometry}},
  author={Chen, Zijie and Xu, Yong and Yuan, Shenghai and Xie, Lihua},
  journal=RAL,
  volume={9},
  number={2},
  pages={1883--1890},
  year={2024}
}

@article{zheng2024fastlivo2,
  title={{Fast-LIVO2: Fast, direct lidar-inertial-visual odometry}},
  author={Zheng, Chunran and Xu, Wei and Zou, Zuhao and Hua, Tong and Yuan, Chongjian and He, Dongjiao and Zhou, Bingyang and Liu, Zheng and Lin, Jiarong and Zhu, Fangcheng and others},
  journal=TRO,
  year={2024},
}

@inproceedings{brachmann2024scene,
  title={{Scene coordinate reconstruction: Posing of image collections via incremental learning of a relocalizer}},
  author={Brachmann, Eric and Wynn, Jamie and Chen, Shuai and Cavallari, Tommaso and Monszpart, Aron and Turmukhambetov, Daniyar and Prisacariu, Victor Adrian},
  booktitle=ECCV,
  pages={421--440},
  year={2024},
}

@article{herau2025pose,
  title={{Pose Optimization for Autonomous Driving Datasets using Neural Rendering Models}},
  author={Herau, Quentin and Piasco, Nathan and Bennehar, Moussab and Rold{\~a}o, Luis and Tsishkou, Dzmitry and Liu, Bingbing and Migniot, Cyrille and Vasseur, Pascal and Demonceaux, C{\'e}dric},
  journal={arXiv preprint arXiv:2504.15776},
  year={2025}
}

@article{xie2021segformer,
  title={{SegFormer: Simple and efficient design for semantic segmentation with transformers}},
  author={Xie, Enze and Wang, Wenhai and Yu, Zhiding and Anandkumar, Anima and Alvarez, Jose M and Luo, Ping},
  journal=NIPS,
  volume={34},
  pages={12077--12090},
  year={2021}
}
}

\end{document}